\documentclass{ecai2012}
\usepackage{latexsym}
\usepackage{times}
\usepackage{helvet}
\usepackage{courier}
\usepackage{graphicx}
\usepackage{algorithmic}
\usepackage{algorithm}
\usepackage{hhline}
\usepackage{multirow}
\usepackage[compact]{titlesec}

\newcommand{\openGoals}{openGoals}
\newcommand{\view}{view}

\newcommand{\G}{{\mathcal G}}

\newcommand{\D}{{\mathcal D}}
\newcommand{\F}{{\mathcal F}}
\newcommand{\I}{{\mathcal I}}
\newcommand{\A}{{\mathcal A}}
\newcommand{\T}{{\mathcal T}}

\newcommand{\AG}{{\mathcal{AG}}}

\newcommand{\Prop}{{\mathcal V}}

\newcommand{\Obj}{{\mathcal{O}}}

\begin{document}

\title{An approach to multi-agent planning with incomplete information}
\author{Alejandro Torre\~no, Eva Onaindia, \' Oscar Sapena \\
Universitat Polit\`ecnica de Val\`encia\\
Camino de Vera s/n\\
46022 Valencia, SPAIN\\
}

\author{Alejandro Torre\~no \and Eva Onaindia \and \' Oscar Sapena\institute{Universitat Polit\`ecnica de Val\`encia, Camino de Vera s/n, 46022 Valencia, Spain} }

\maketitle

\begin{abstract}
Multi-agent planning (MAP) approaches have been typically conceived for independent or loosely-coupled problems to enhance the benefits of distributed planning between autonomous agents as solving this type of problems require less coordination between the agents' sub-plans. However, when it comes to tightly-coupled agents' tasks, MAP has been relegated in favour of centralized approaches and little work has been done in this direction. In this paper, we present a general-purpose MAP capable to efficiently handle planning problems with any level of coupling between agents. We propose a cooperative refinement planning approach, built upon the partial-order planning paradigm, that allows agents to work with incomplete information and to have incomplete views of the world, i.e. being ignorant of other agents' information, as well as maintaining their own private information. We show various experiments to compare the performance of our system with a distributed CSP-based MAP approach over a suite of problems.
\end{abstract}

\section{INTRODUCTION}

Multi-agent planning (MAP) refers to any planning or plan execution activity that involves several agents. In general terms, MAP is about the collective effort of multiple planning agents to combine their knowledge, information, and capabilities so as to develop solutions to problems that each could not have solved as well (if at all) alone \cite{Durfee01a}. There exists a great variety of tools and techniques for MAP. Agent-oriented MAP approaches put the emphasis on distributed execution, plan synchronization and collaborative activity at run-time planning to ensure that the agent's local objectives will be met \cite{Desjardins99a,Tambe97}. Another research line in MAP focuses on coordination of already completed plans that agents have constructed to achieve their individual goals, as for example plan merging \cite{ToninoBWW02,CoxDB05,CoxD05}. In contrast, the cooperative distributed planning (CDP) approach puts the emphasis on planning and how it can be extended into a distributed environment, on building a competent plan carried out by multiple agents \cite{Desjardins99a}. In CDP, agents typically exchange information about their plans, which they iteratively refine and revise until they fit together well.

Following the cooperative approach, differences among MAP models lie in the integration of the planning and coordination stages \cite{Durfee01a,deWeerdt09}. Some recent works on fully cooperative MAP have emerged lately. The work in \cite{Kvarnstrom11} considers agents as having sequential threads of execution and interaction only occurs when distributing sub-plans to individual agents for plan execution. This approach follows a single-agent planning and distributed coordination. A centralized algorithm for MAP can be found in \cite{Brafman08}, where multiple agents do planning over a centralized plan interleaving planning and coordination. In a distributed version of this latter work, authors use a distributed CSP solver to handle coordination \cite{Nissim10}.

The aforementioned approaches are conceived for loosely-coupled problems (LCP), where agents have little interaction between each other, as these processes are likely to be inefficient in tightly-coupled problems (TCP) \cite{Nissim10}. This way, the coupling level of a cooperative multi-agent system is formally defined as a set of parameters to limit the combinatorial blow-up of planning complexity \cite{Brafman08}. On the other hand, these MAP models do not consider systems composed of multiple entities distributed functionally or spatially but rather agents endowed with the same capabilities and acting under complete information. When capabilities are distributed across the agents' domains, agents have \emph{necessarily} to interact to solve the MAP problem while being unaware of the other agents' abilities or information about the world, i.e. working under incomplete information.

In this paper, we present a general-purpose MAP model able to work with inherently distributed entities and suitable for both LCP and TCP domains. Similarly to \cite{JonssonR11}, we use an iterative planning refinement procedure that uses single-agent planning technology. Particularly, our model builds upon a partial-order planning (POP) paradigm, which also allow us to represent a collection of acting entities as a single agent. POP is a very suitable approach for centralized MAP with a small number of coordination points between agents \cite{Kvarnstrom11}, and the application of a multi-agent POP refinement framework also reveals as a very appropriate mechanism to address tightly-coupled problems.

This paper is organized as follows. The next section presents the specification of a MAP task. Following, we explain the POP refinement approach and the extensions we have introduced to deal with a multi-agent representation and incomplete information. The next sections describe our MAP task theoretical model and the refinement planning algorithm carried out by the agents. Following, we show the results of the tests we have performed, and finally, we conclude and outline the future lines of research.

\section{MULTI-AGENT PLANNING TASK}

In our approach, the planning formalism of an agent is based on a STRIPS-like model of classical planning under partial observability. The model allows agents to represent their partial view of the world through the adoption of the open world assumption. States are represented in terms of state variables. $\Obj$ is a finite set of objects that model the elements of the planning domain; $\Prop$ is a finite set of \emph{state variables} each with an associated finite domain, $\D_v$, of mutually exclusive values. Values in $\D_v$ denote objects of the planning domain,  i.e., $\forall v \in \Prop$, $\D_v \subseteq \Obj$. A state is a set of \emph{positive fluents} of the form $\langle v, d \rangle$, and \emph{negative fluents} of the form $\langle v, \neg d \rangle$, meaning that the variable takes on the value $d$ or $\neg d$, respectively. A \emph{formula} $(v, d)$ evaluates to true if the fluent $\langle v, d \rangle$ is present in the state and it evaluates to false otherwise. More specifically, $(v, d)$ evaluates to false if the fluent $\langle v, \neg d \rangle$ is in the state, or if no fluent relating the variable, $v$, and the value, $d$, is present in the state, in which case we say the current value of $v$ is unknown. We will generally refer to as \emph{fluents} both positive and negative fluents.

\emph{Actions} are given as tuples $a= \langle {\sf pre}(a), {\sf eff}(a) \rangle$, where ${\sf pre}(a)$ denotes the formulas that must hold in a state $S$ for $a$ to be applicable, and ${\sf eff}(a)$ represents the new fluents in the resulting state $S'$. Effects of the form $(v = d)$ add a fluent $\langle v, d\rangle$ in the resulting state as well as a set of fluents $\{\langle v, \neg d_j\rangle\}, \forall d_j \neq d, d_j \in \D_v$, reflecting that $(v, d_j)$ evaluates to false in the resulting state. Effects of the form $(v \neq d)$ add a fluent $\langle v, \neg d\rangle$ to the resulting state, which implies the current value of $v$ is unknown unless there is a fluent $\langle v, d'\rangle$ in $S'$, $d \neq d'$.

\vspace{0,2cm}
We define a MAP task as a tuple $\T = \langle\AG, \Prop, \A, \I, \G\rangle$ where:
\vspace{-0.2cm}
\begin{itemize}
\item $\AG=\{1, \ldots, n\}$ is a finite non-empty set of planning agents.
 \item $\Prop=\{\Prop_i\}_{i=1}^n$, where $\Prop_i$ is the set of state variables managed by agent $i$. Variables can be shared by two or more different agents.
 \item $\A=\{\A_i\}_{i=1}^n$, where $\A_i$ is the set of actions that agent $i$ can perform. Given two different agents $i$, $j$, $\A_i$ and $A_j$ can share some common actions or be two disjoint sets.
 \item $\I=\{\I_i\}_{i=1}^n$, where $\I_i$ is the set of fluents known by agent $i$ that represents the \emph{initial state} of the agent. If two agents share a variable $v$ then they also share all of the fluents regarding $v$.
  \item $\G=\{\G_i\}_{i=1}^n$, where $\G_i$ is a set of formulas known to agent $i$ that must hold in the final state and denote the top-level goals of $\T$.
\end{itemize}
\vspace{-0.1cm}
As defined above, state variables may not be known to all agents. Given a state variable $v \in \Prop_i$ and $v \not \in \Prop_j$, $\forall j \neq i$, $v$ is said to be \emph{private} to agent $i$. Additionally, agents can have different visions of the domain of a state variable; that is, not every value in a variable domain has to be visible to all agents. Given an agent $i$, we denote its view of the domain of a variable $v$ as $\D_{v_i} \subseteq \D_v$. Thus, the domain of a state variable $v$ can be defined as $\D_v=\{{\D_{v_i}\}}_{i=1}^n$. Agents' incomplete views on the state variables and their domains directly affect the visibility of the fluents.
\vspace{-0.2cm}
\begin{itemize}
	\item An agent $i$ has \emph{full visibility} of a fluent $\langle v, d \rangle$ or $\langle v, \neg d \rangle$ if $v \in \Prop_i$ and $d \in \D_{v_i}$.
	\item An agent $i$ has \emph{partial visibility} of a fluent $\langle v, d \rangle$ or $\langle v, \neg d \rangle$ if $v \in \Prop_i$ but $d \not \in \D_{v_i}$. Given a state $S$, where $\langle v, d \rangle \in S$, agent $i$ will see instead a fluent $\langle v, \perp \rangle$, where $\perp$ is the undefined value.
	\item An agent $i$ has \emph{no visibility} of a fluent $\langle v, d \rangle$ or $\langle v, \neg d \rangle$ if $v \not \in \Prop_i$.
\end{itemize}
\vspace{-0.1cm}

Our MAP model can be viewed as a POP-based, multi-agent refinement planning framework, a general method based on the refinement of the set of all possible partial-order plans \cite{Kambhampati97}. An agent proposes a plan $\Pi$ that typically enforces some top-level goals of the planning task; then, the rest of agents collaborate on the refinement of this base plan $\Pi$ by proposing refinement steps that solve some \emph{open goals} in ${\sf\openGoals}(\Pi)$. This way, agents cooperatively solve the MAP task by consecutively refining an initially empty plan $\Pi$.

A \emph{refinement step} $\Pi_i$ devised by an agent $i$ over a base plan $\Pi^{g}$, where $g \in {\sf\openGoals}(\Pi^{g})$, is a triple $\Pi_i = \langle \Delta, OR, CL \rangle$, where $\Delta \in \A_i$ is a set of actions and $OR$ and $CL$ are sets of \emph{orderings} and \emph{causal links} over $\Delta$, respectively. $\Pi_i$ is a plan free of \emph{threats} \cite{Younes03} that solves $g$ as well as all the new open goals that arise from this resolution and can only be achieved by agent $i$, $\langle v, d\rangle$ or $\langle v, \neg d\rangle$, where $(v \in \Prop_i) \wedge (v \not\in \Prop_j, \forall j \neq i)$. That is, when solving an open goal of a base plan, an agent $i$ will also achieve the new arising open goals concerning fluents that are only visible to $i$, so are not visible to the rest of agents, leaving the rest of goals unsolved. Let $g \in {\sf\openGoals}(\Pi^{g})$ be a formula of the form $(v, d)$ or $(v, \neg d)$; an agent $i$ computes a refinement step over $\Pi^{g}$ iff $v \in \Prop_i$.

Plans that agents build are concurrent multi-agent (MA) plans as two different actions in $\Pi$ can now be executed concurrently by two different agents. Some MAP models adopt a simple form of concurrency: two actions can happen simultaneously if none of them changes the value of a state variable that the other relies on or affects, too \cite{Brenner09}. We impose the additional concurrency constraint that the preconditions of two actions have to be mutually consistent \cite{BoutilierB01}. This definition of concurrency is straightforwardly extended to a joint action $a=\langle a_1, \ldots, a_n\rangle$. Agents address concurrency inconsistencies through the detection of threats over the causal links of their actions. This way, concurrency issues between two different actions may not arise until their preconditions are supported through causal links.

A \emph{refinement plan} $\Pi$ devised by an agent $i$ over a base plan $\Pi^{g}$ is a concurrent MA plan that results from the composition of $\Pi^{g}$ and a refinement step $\Pi_i$ proposed by agent $i$. This refinement plan, which could eventually become a base plan, is defined as $\Pi=\Pi^{g} \circ \Pi_i$, where $\circ$ represents the composition operation. A composite plan $\Pi$ is a concurrent MA plan if for every pair of unequal actions $a_i$ and  $a_j$, $i \not= j$, $\forall p_i \in {\sf pre}(a_i), p_i \not \in {\sf\openGoals}(\Pi)$, $\forall p_j \in {\sf pre}(a_j), p_j \not \in {\sf\openGoals}(\Pi)$, $a_i$ and $a_j$ are concurrently consistent.

In our model, each agent implements a POP planner to compute refinement plans over a base plan $\Pi$. If an agent is not capable to come up with a concurrent MA plan, then the agent refrains from suggesting such a refinement. If no agent elicits a consistent refinement plan for a base plan, the plan node is pruned.

\vspace{-0.1cm}
\begin{algorithm}
\caption{Dis-RPG construction for an agent $i$}
\label{RPG_algorithm}
\begin{algorithmic}
\STATE Build initial $RPG_i$
\REPEAT
	\STATE $\forall j \neq i$, $i$ sends $j$ shareable fluents $SF_{i\rightarrow j} \in RPG_i$ of the form $\langle v, d\rangle$ or $\langle v, \neg d\rangle$, where $v \in \Prop_i \cap \Prop_j$ and $d \in \D_{v_i} \cap \D_{v_j}$
	\STATE $\forall j \neq i$, $i$ receives from $j$ shareable fluents $SF_{j\rightarrow i} \in RPG_j$ of the form $\langle v, d\rangle$ or $\langle v, \neg d\rangle$, where $v \in \Prop_i \cap \Prop_j$ and $d \in \D_{v_i} \cap \D_{v_j}$
	\STATE $RF \gets \emptyset$
	\STATE $\forall j \neq i, RF_i \gets RF_i \cup SF_{j\rightarrow i}$
	\FORALL{received fluents $f \in RF_i$}
	\IF {$f \not\in RPG_i$}
		\STATE Insert f in $RPG_i$
		\STATE $cost_{RPG_i}(f) \gets cost(f)$
	\ENDIF
	\IF {$(f \in RPG_i)$ $\wedge$ $(cost_{RPG_i}(f) > cost(f))$}
		\STATE $cost_{RPG_i}(f) \gets cost(f)$
	\ENDIF
	\ENDFOR
	\STATE Expand $RPG_i$
\UNTIL $RF_i = \emptyset$
\end{algorithmic}
\end{algorithm}
\vspace{-0.2cm}

\section{REFINEMENT PLANNING}
\label{Refinement}

The cooperative refinement planning algorithm starts with a preliminary information exchange by which agents communicate shareable information. After this initial stage, agents execute the multi-agent refinement planning algorithm, which comprises two interleaved stages. First, agents individually elicit refinement plans over a centralized base plan through their embedded POP. Later, agents jointly select the most promising refinement as the next base plan.

\subsection{Information exchange}
Agents receive the information on the MAP task through a set of definition files. These files are encoded in a MAP language that extends \emph{PDDL3.1} \cite{Kovacs11}, including a {\ttfamily :shared-data} section to configure the agent's vision of the planning task and which fluents it shares and with whom.

Prior to executing the refinement procedure, agents share information by building a distributed Relaxed Planning Graph (dis-RPG), based on the approach of \cite{Feng07}. Agents  exchange the fluents defined as shareable in the {\ttfamily :shared-data} section of the MAP definition files. Fluents are labeled with the list of agents that can achieve them, giving each agent a view of the possible interactions that can arise at planning time with other agents. Additionally, the dis-RPG provides an estimate of the best cost to achieve each fluent, a helpful information to design heuristics to guide the problem-solving process.

Algorithm \ref{RPG_algorithm} summarizes the construction of the dis-RPG. Agents compute an initial RPG and expand it by following the procedure in \cite{Hoffmann01}. The RPG contains a set of fluent and action levels that are interleaved. The first fluent level contains the fluents that are part of the initial state, and the first action level includes all the actions whose preconditions appear in the first fluent level. The effects of these actions are placed in the second fluent level, and this way the graph is expanded until no new fluents are found.

Once all the agents have computed their initial RPGs, the iterative dis-RPG composition begins. As depicted in Algorithm \ref{RPG_algorithm}, agents start each iteration by exchanging the the fluents shareable with other agents. An agent $i$ will send agent $j$ the set of fluents $SF_{i\rightarrow j}$ that are visible to agent $j$, i.e., the new fluents of the form $\langle v, d\rangle$ or $\langle v, \neg d\rangle$, where $v \in \Prop_i \cap \Prop_j$ and $d \in \D_{v_i} \cap \D_{v_j}$. Likewise, agent $i$ will receive from all agents $j \neq i$ the shareable fluents they have generated.

Agent $i$ updates then its $RPG_i$ with the set of new fluents it has received, $RF_i$. If a fluent $f$ is not yet in $RPG_i$, it is stored according to $cost(f)$. If $f$ is already in $RPG_i$, its cost is updated if $cost_{RPG_i}(f) > cost(f)$. Hence, agents only store the best estimated cost to reach each fluent. After updating $RPG_i$, agent $i$ expands it by checking whether the new inserted fluents trigger new actions in $RPG_i$ or not. The fluents produced as effects of these new actions will be shared in the next iteration.

The process finishes when there are no new fluents in the system. Following, agents start the refinement planning process to build a solution plan jointly.

\subsection{Multi-agent refinement planning}

The refinement planning process is based on a democratic leadership by which a baton is scheduled among the agents following a round-robin strategy. Agents carry out two interleaved stages: the individual construction of refinement plans through a POP, and a coordination process by which agents jointly search the refinement space.

Algorithm \ref{Problem_solving_algorithm} describes the refinement planning process. Each agent $i$ computes a finite set of refinement plans for $\Pi^g$, $Refinements_i (\Pi^g)$, through its embedded POP planner. The internal POP system follows an A$^*$ search algorithm guided by a state-of-the-art POP heuristic function  \cite{Younes03}. The resulting refinement plans are exchanged by the agents in the system for their evaluation (send and receive operations in Algorithm \ref{Problem_solving_algorithm}).

Agent $i$ has a local, partial vision of each refinement plan, $\view_i(\Pi)$, according to its visibility over the planning task $\T$. Thus, when receiving a refinement plan $\Pi$, agent $i$ will only view the open goals $(v, d)\in {\sf\openGoals}(\Pi)\;|\;v \in \Prop_i$. With respect to the fluents, agent $i$ will only view those fluents for which it has \emph{full visibility}. If $i$ has \emph{partial visibility} of a fluent $\langle v, d \rangle$ or $\langle v, \neg d \rangle$, it will see instead a fluent $\langle v, \perp \rangle$, where $\perp$ stands for the undefined value. This notion of partial view directly affects the evaluation of the refinements.

The evaluation of refinement plans is carried out through a utility function $\F$ (currently, we use the same heuristic function that guides the agents' internal POP for this purpose) that allows agents to estimate the quality of the plans. Since agents do not have complete information on the MAP task or the refinement plans, they evaluate plans according to its own view of each refinement plan $\Pi$, i.e., agent $i$ evaluates a refinement plan $\Pi$ according to $\F(\view_i(\Pi))$ (see Algorithm \ref{Problem_solving_algorithm}).

\vspace{-0.2cm}
\begin{algorithm}
\caption{Refinement planning process for an agent $i$}
\label{Problem_solving_algorithm}
\begin{algorithmic}
\STATE $\Pi \gets \Pi_0$
\STATE $R = \emptyset$
\REPEAT
	\STATE Select open goal $g \in {\sf\openGoals} (\Pi)$
	\STATE Refine base plan $\Pi^{g}$ individually
    \STATE $\forall j \not= i$, send $Refinements_i(\Pi^{g})$ to agent $j$
    \STATE $\forall j \not= i$, receive $Refinements_j(\Pi^{g})$
    \STATE $Refinements(\Pi^{g}) \gets Refinements_i(\Pi^{g})$
    \STATE $\forall j \not= i$, $Refinements(\Pi^{g}) \gets Refinements(\Pi^{g}) \cup$ 
	\STATE $Refinements_j(\Pi^{g})$
	\FORALL{plans $\Pi \in Refinements(\Pi^{g})$}
		\STATE Evaluate $\Pi$ according to $\F(\view_i(\Pi))$
	\ENDFOR
	\STATE $R \gets R \cup Refinements(\Pi^{g})$
	\STATE Select best-valued plan $\Pi_{best} \in R$
	\STATE $\Pi \gets \Pi_{best}$
	\IF {${\sf \openGoals}(\Pi) = \emptyset$}
		\RETURN $\Pi$
	\ENDIF
\UNTIL $R = \emptyset$
\end{algorithmic}
\end{algorithm}
\vspace{-0.2cm}

Once evaluated, the new refinement plans are stored in the set of refinements $R$. Next, each agent votes for the best-valued candidate $\Pi_{best} \in R$. In case of a draw, the baton agent will choose the next base plan among the most voted alternatives.

Once a refinement plan is selected, agents adopt it as the new base plan $\Pi$. If ${\sf\openGoals}(\Pi) = \emptyset$, a solution plan is returned. As some open goals might not be visible to some agents, every agent $i$ must confirm that $\Pi$ is a solution plan according to $\view_i(\Pi)$, i.e., $\Pi$ is a solution iff $\forall i \in \AG, {\sf\openGoals}(view_i(\Pi)) = \emptyset$. If the plan has still pending goals, the baton agent selects the next open goal $g \in {\sf\openGoals}(\Pi)$ to be solved, and a new iteration of the refinement planning process starts.

The planning algorithm carried out by the agents can be regarded as a joint exploration of the refinement space. Nodes in the search tree represent refinement plans and each iteration of the algorithm expands a different node.

\subsection{Soundness and completeness}

The algorithm we have presented can be regarded as a multi-agent extension of the POP algorithm. A partial-order plan is sound if it is a threat-free plan. In our algorithm, we address inconsistencies among the concurrent MA plans by detecting and solving threats. Thus, in order to prove that our algorithm is sound, we should ensure that all the threats among the causal links of a concurrent MA plan are correctly detected and solved.

Under complete information, threats on a MA concurrent plan will be correctly detected by any agent, as all the fluents in the plan are fully visible. In our incomplete information model, we should study how visibility over fluents affects the detection of threats.

Let $\Pi$ be a MA concurrent plan and let $\langle v, d_1 \rangle$ be a fluent in a causal link $cl \in CL(\Pi)$. Suppose that an agent $i$ builds a refinement $\Pi'$ over $\Pi$ that adds a new action $a_t$ to the plan which is not ordered with respect to $cl$ and has an effect $(v = d_2)$. This effect causes a threat over $cl$ as it conflicts with $\langle v, d_1 \rangle$. For $\Pi'$ to be sound, agent $i$ should be able to detect such a threat whatever visibility it has over the fluent $\langle v, d_1 \rangle$:
\vspace{-0.2cm}
\begin{itemize}
	\item If $i$ has \emph{full visibility} over $\langle v, d_1 \rangle$, the inconsistency between $cl$ and $a_t$ will be correctly detected.
	\item If $i$ has \emph{no visibility} over $\langle v, d_1 \rangle$, then $v \not \in \Prop_i$. In this case, agent $i$ does not have an action $a_t$ with an effect involving variable $v$, i.e., such a threat can never occur.
	\item If $i$ has \emph{partial visibility} over $\langle v, d_1 \rangle$, agent $i$ will see instead a fluent $\langle v, \perp \rangle$. Since $\perp \neq d_2$, the threat will be detected and solved.
\end{itemize}
\vspace{-0.1cm}
Therefore, all the threats over MA concurrent plans are always detected and resolved, which proves that our MAP algorithm is sound.

As for completeness, we cannot ensure that our MAP algorithm is complete. According to the notion of refinement plan we have used in this work, the number of refinement plans that an agent can produce over a base plan may not be finite. Hence, we are implicitly pruning the refinement search space. Nevertheless, agents rely on an A$^*$ POP search process to build the refinement plans, which in most cases returns good refinement plans that guide the MAP algorithm towards a solution plan. The empirical results shown in the next section confirm our claim.

\vspace{-0.4cm}
\begin{figure}
\centering
\includegraphics[width=8.1cm]{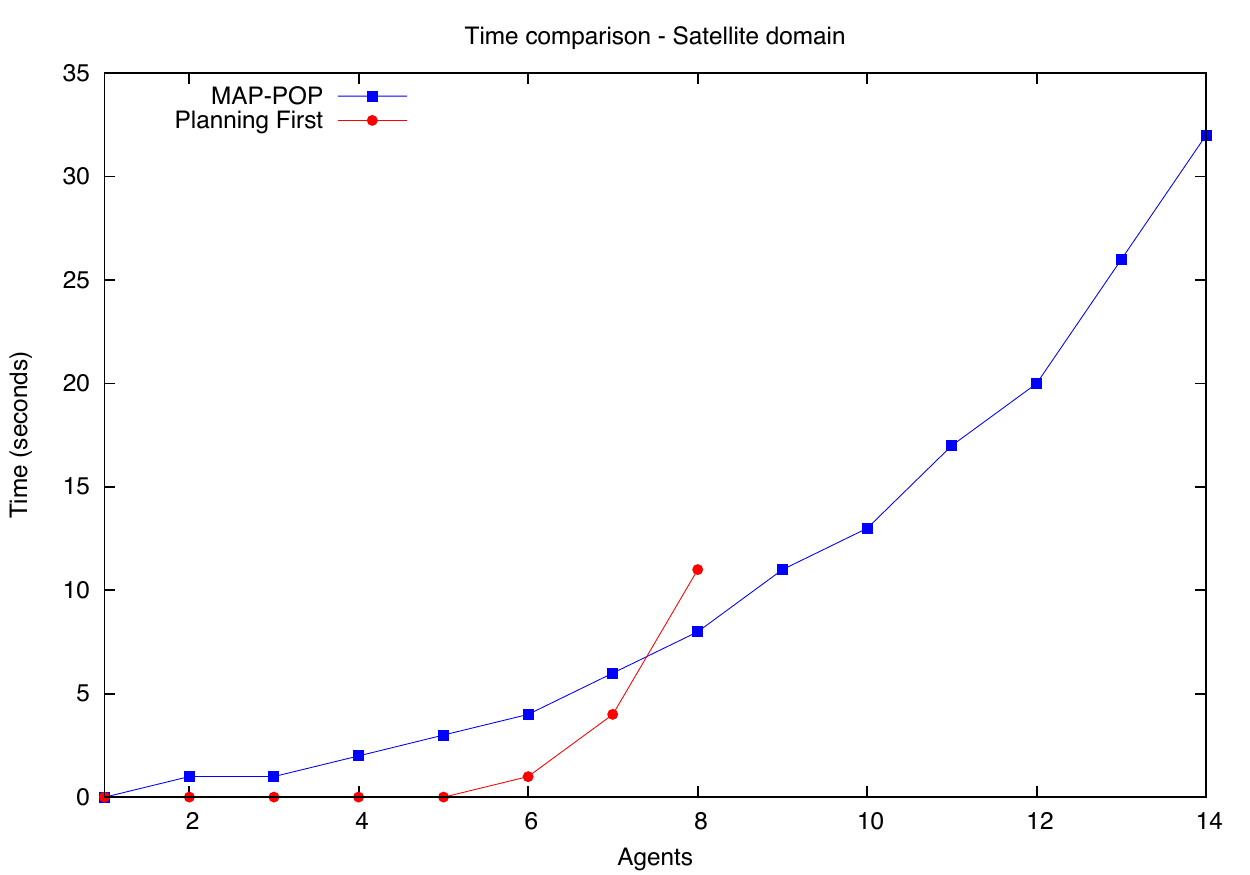}
\vspace{-0.4cm}
\caption{Scalability results for the Satellite domain}
\label{GraphSatellite}
\end{figure}
\vspace{-0.4cm}

\section{EXPERIMENTAL RESULTS}

We designed and executed a set of tests to compare the performance and scalability of our MAP-POP approach with another state-of-the-art MAP system. Comparing the performance of multi-agent planning systems is not an easy task due to two main reasons. First, most MAP approaches are not general-purpose but domain-dependent systems specifically designed to address a particular problem, most typically traffic control or real-time planning applications. Second, unlike single-agent planners that have been promoted and populated through the celebration of the International Planning Competitions\footnote{http://ipc.icaps-conference.org/} (IPC) and, therefore, have been made publicly available, it is difficult to find a multi-agent planner able to run the benchmark domains and planning problem suites created for the IPCs.

Despite these drawbacks, we could assess the performance of MAP-POP and compare the results with those obtained in the Planning First approach presented in \cite{Nissim10}\footnote{We want to especially thank Raz Nissim for providing us with the source code of his Planning First system for testing and comparison purposes.}. Planning First is a MAP system that also makes use of single-agent planning technology. More precisely, it builds upon a single-agent state-based planner \cite{Coles08}, and handles agent coordination by solving a distributed CSP.

Planning First defines public actions as the actions of an agent whose descriptions contain atoms affected by and/or affecting the actions of another agent. Based on this concept, it defines the notion of coupling level as the average rate of public actions of an agent. A high value of coupling level results in many agent coordination points, thus giving rise to tightly-coupled problems. The approach followed by Planning First is especially effective when dealing with loosely-coupled problems (LCP) \cite{Nissim10}, but its performance decreases when tackling tightly-coupled problems (TCP).

\vspace{-0.3cm}
\begin{figure}
\centering
\includegraphics[width=8.1cm]{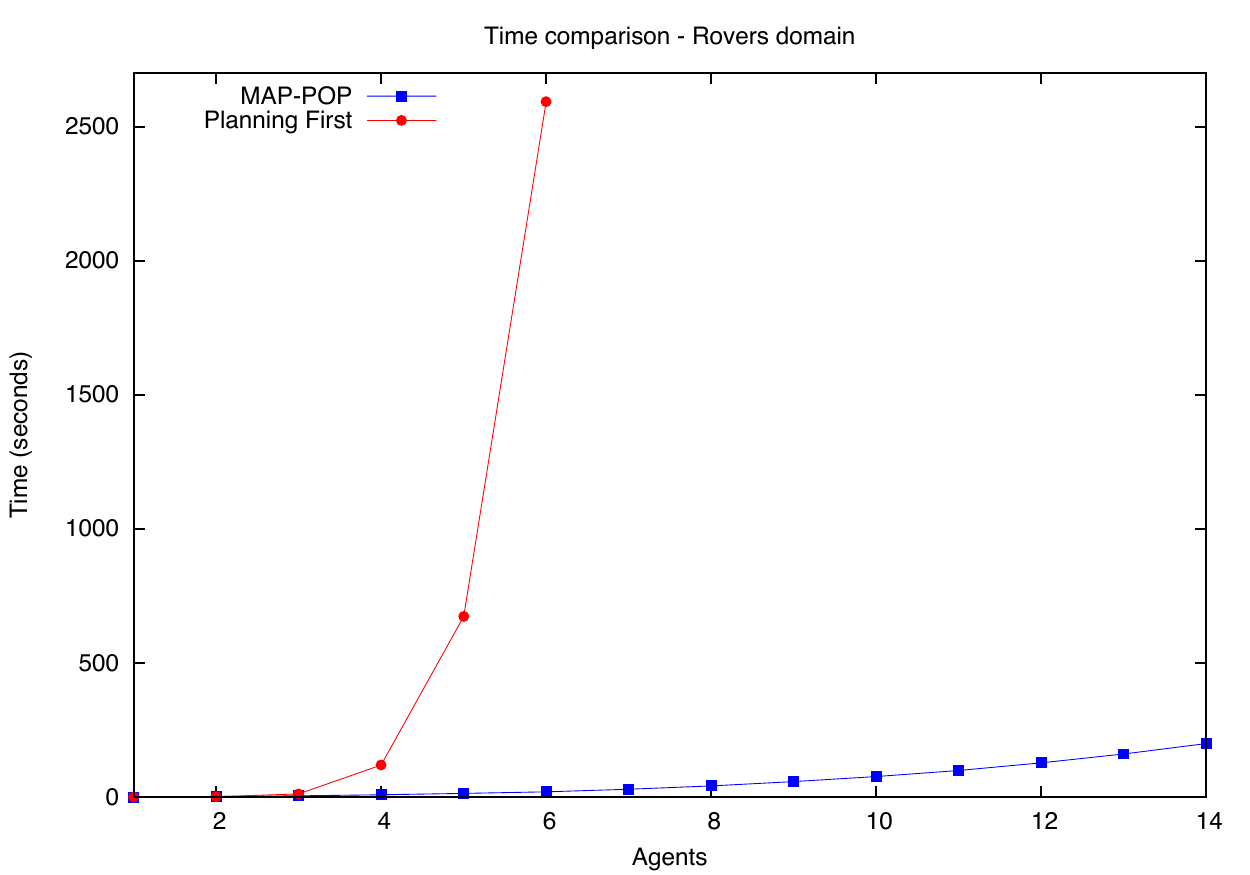}
\vspace{-0.4cm}
\caption{Scalability results for the Rovers domain}
\label{GraphRovers}
\end{figure}
\vspace{-0.4cm}

The tests presented here involve three of the benchmark domains used on the IPCs: \emph{satellite}, \emph{rovers} and \emph{logistics}, which are the domains used in the results presented in \cite{Nissim10} as well. These domains give rise to problems of different coupling levels. The \emph{satellite} problems are LCP as the different agents (the satellites) are not likely to interact with each other; they move, calibrate their instruments and take images by themselves. \emph{Rovers} problems tend to present a medium coupling level: rover agents are independent but they have access to certain shared resources in their environment, namely the rock and soil samples they collect and analyze. The \emph{logistics} problems fall into the TCP category since agents (trucks and planes) have to cooperate to transport the different packages to the target locations and problems present several coordination points (locations) at which agents can interact.

We adapted the STRIPS problem files used in the IPCs to both our MAP language and Nissim's \emph{MA-STRIPS} language. Problems from the IPCs turned out to be complex instances for Planning First because agents have necessarily to interact to each other and cooperate to find a solution plan for these problems and Planning First works better when plans for each agent can be computed (mostly) independently. For this reason, we encoded an additional set of problems limiting cooperation and interactions among agents as much as possible. Particularly, in these additional problems, agents can solve goals independently, i.e., an agent is able to solve a goal or set of goals by itself without need of interacting with the rest of agents (we will refer to these problems as independent problems in the remainder).

\begin{center}
\begin{table*}
\centering
{\footnotesize
\begin{tabular}{ | c | c | c | c || r | r | r | r || r | r | r | r |}
	\hhline{------------}
    \multicolumn{1}{|c|}{\multirow{2}{*}{Problem}} &
	\multicolumn{1}{c|}{\multirow{2}{*}{$\#$Agents}} &
    \multicolumn{1}{c|}{\%Coupling} &
    \multicolumn{1}{c||}{$\#$Domain} &
    \multicolumn{4}{c||}{MAP-POP} &
    \multicolumn{4}{c|}{Planning First}\\ \cline{5-12}
    \multicolumn{1}{|c|}{} &
    \multicolumn{1}{c|}{} &
    \multicolumn{1}{c|}{level} &
    \multicolumn{1}{c||}{actions} &
    \multicolumn{1}{c|}{$\#$Acts} &
    \multicolumn{1}{c|}{$\#$TS} &
    \multicolumn{1}{c|}{$\#$Partics} &
    \multicolumn{1}{c||}{Time} &
    \multicolumn{1}{c|}{$\#$Acts} &
    \multicolumn{1}{c|}{$\#$TS}&
    \multicolumn{1}{c|}{$\#$Partics} &
    \multicolumn{1}{c|}{Time} \\ \hhline{============}
    \multicolumn{1}{|c|}{IPCSat1} & 1 & 1,2 & 54 & 9 & 8 & 1 & 0,23 & 10 & 9 & 1 & 0,14 \\ \hline
    \multicolumn{1}{|c|}{IPCSat4} & 2 & 29,3 & 2082 & 21 & 11 & 2 & 18,80 &  & & & $\dagger$ \\ \hline
    \multicolumn{1}{|c|}{IPCSat10} & 5 & 23,7 & 1786 & 29 & 20 & 3 & 90,3 &  & & & $\dagger$ \\ \hline
    \multicolumn{1}{|c|}{IPCSat16} & 10 & 18,3 & 7196 & 51 & 24 & 5 & 73,7 &  & & & $\dagger$ \\ \hline
    \multicolumn{1}{|c|}{IPCSat17} & 12 & 14,3 & 8324 & 46 & 16 & 4 & 53,9 & & & & $\dagger$\\ \hhline{============}

    \multicolumn{1}{|c|}{IndSat1} & 2 & 5,2 & 40 & 9 & 4 & 2 & 0,83 & 9 & 4 & 2 & 0,16 \\ \hline
    \multicolumn{1}{|c|}{IndSat2} & 4 & 1,4 & 274 & 14 & 3 & 4 & 2,20 & 14 & 4 & 4 & 0,31 \\ \hline
    \multicolumn{1}{|c|}{IndSat3} & 7 & 0,3 & 1820 & 32 & 4 & 7 & 6,5 & 32 & 4 & 7 & 4,1 \\ \hline
    \multicolumn{1}{|c|}{IndSat4} & 8 & 0,3 & 2082 & 28 & 3 & 8 & 8,7 & 28 & 4 & 8 & 11,1 \\ \hline
    \multicolumn{1}{|c|}{IndSat5} & 14 & 0,1 & 11020 & 63 & 4 & 14 & 32,5 & & & & $\dagger$ \\ \hhline{============}

	\multicolumn{1}{|c|}{IPCRov1} & 1 & 1,2 & 81 & 10 & 7 & 1 & 0,344 & 11 & 7 & 1 & 0,359 \\ \hline
    \multicolumn{1}{|c|}{IPCRov2} & 1 & 2,3 & 45 & 8 & 4 & 1 & 0,390 & 9 & 5 & 1 & 0,312 \\ \hline

    \multicolumn{1}{|c|}{IPCRov7} & 3 & 77,4 & 157 & 18 & 6 & 3 & 8,578 & & & & $\dagger$\\ \hline
    \multicolumn{1}{|c|}{IPCRov14} & 4 & 58,7 & 797 & 35 & 21 & 2 & 81,874 & & & & $\dagger$\\ \hline
    \multicolumn{1}{|c|}{IPCRov15} & 4 & 85 & 536 & 42 & 16 & 4 & 42,014 & & & & $\dagger$\\ \hhline{============}
    \multicolumn{1}{|c|}{IndRov1} & 2 & 45,5 & 160 & 24 & 11 & 2 & 3,609 & 22 & 7 & 2 & 2,75\\ \hline
    \multicolumn{1}{|c|}{IndRov2} & 3 & 45,5 & 239 & 36 & 11 & 3 & 5,500 & 33 & 7 & 3 & 12,141\\ \hline
    \multicolumn{1}{|c|}{IndRov3} & 4 & 45,5 & 318 & 48 & 11 & 4 & 9,188 & 44 & 7 & 4 & 120,719\\ \hline
    \multicolumn{1}{|c|}{IndRov4} & 5 & 45,5 & 397 & 70 & 11 & 5 & 14,141 & 55 & 7 & 5 & 674\\ \hline
    \multicolumn{1}{|c|}{IndRov5} & 6 & 45,5 & 476 & 72 & 11 & 6 & 20,688 & 66 & 7 & 6 &  2594,515\\ \hhline{============}

    \multicolumn{1}{|c|}{IPCLog2} & 3 & 20 & 52 & 27 & 9 & 3 & 18,187 & & & & $\dagger$\\ \hline
    \multicolumn{1}{|c|}{IPCLog4} & 4 & 12,3 & 116 & 37 & 13 & 4 & 33,765  & & & & $\dagger$\\ \hline
    \multicolumn{1}{|c|}{IPCLog5} & 4 & 14 & 116 & 31 & 11 & 4 & 40,188  & & & & $\dagger$\\ \hline
    \multicolumn{1}{|c|}{IPCLog7} & 5 & 9,8 & 206 & 46 & 15 & 5 & 96,484 & & & & $\dagger$\\ \hline
    \multicolumn{1}{|c|}{IPCLog9} & 5 & 11,7 & 206 & 45 & 17 & 5 & 239,578  & & & & $\dagger$\\ \hhline{============}
    \multicolumn{1}{|c|}{IndLog1} & 3 & 44,4 & 20 & 6 & 6 & 2 & 1,579 & 9 & 8 & 3 & 0,578\\ \hline
    \multicolumn{1}{|c|}{IndLog2} & 3 & 55,5 & 20 & 10 & 9 & 3 & 2,250 & 10 & 9 & 3 & 0,609\\ \hline
    \multicolumn{1}{|c|}{IndLog3} & 4 & 65 & 42 & 13 & 10 & 4 & 3,225 & 9 & 8 & 4 & 66,187 \\ \hline
    \multicolumn{1}{|c|}{IndLog4} & 4 & 70 & 42 & 14 & 6 & 4 & 3,766 & 14 & 6 & 4 & 284,094 \\ \hline
    \multicolumn{1}{|c|}{IndLog5} & 6 & 54,1 & 98 & 21 & 6 & 6 & 13,578 & & & & $\dagger$\\ \hline

\end{tabular}}
\caption{Performance comparison between MAP-POP and Planning First}
\label{Res_MAPSAP}
\end{table*}
\end{center}
\vspace{-0.32cm}

Table \ref{Res_MAPSAP} shows the results when comparing the quality of the solution plans obtained with MAP-POP and Planning First and the execution times\footnote{All the tests were performed on a single machine with a 2.83 GHz Intel Core 2 Quad CPU and 8 GB RAM.}. The quality of the solution plans is assessed through three parameters: a) the number of actions of the plan; b) the duration of the plan, i.e. the number of time units or \emph{time steps} required to execute the plan; and c) the number of agents that take part in the solution plan. This latter parameter gives an idea of how the effort on solving the problem has been distributed among the agents.

Problems labeled with \emph{IPC} are directly taken from the IPC benchmarks, while problems labeled with \emph{Ind} are the extra set of \emph{independent} problems we created to assess Planning First performance (for each domain, we show the results of 5 out of the 20 IPC problems we tested as well as 5 independent problems). The next three columns in the table show the difficulty of the planning problems: \emph{$\#$Agents} indicates the number of agents involved in the problem; \emph{$\%$Coupling level} estimates the coupling level of the problem as the average rate of instantiated public actions of agents (taking into consideration the notion of public and private action defined in \cite{Nissim10}), and \emph{$\#$Domain actions} refers to the total number of instanced actions. The results for each planner include the number of actions(\emph{$\#$Acts}) and time steps (\emph{$\#$TS}) of the solution plan, respectively. \emph{$\#$Partics} indicates the number of agents that take part in the solution plan, and \emph{Time} shows the total execution time. A dagger ($\dagger$) indicates that Planning First was not able to solve the problem.

For the most loosely-coupled problems, the \emph{satellite} domain, MAP-POP exhibited an excellent performance as our results confirmed that it was able to solve 18 out of 20 IPC problems. For the five IPC problems for the \emph{satellite} domain shown in Table \ref{Res_MAPSAP}, we can see that our approach deals very efficiently with complex problems up to 12 agents. It is also noticeable that at least one third of the participating agents take part in the solution plans, which has a positive impact on the plan duration, as actions are carried out in parallel by different agents. Although the IPC \emph{satellite} problems do not present a high coupling level (less than 30$\%$ of public actions in the worst case), Planning First only solves the first IPC problem, as these problems require cooperation among agents and it is more necessary for larger instances. As for the additional problems we encoded (IndSat1, ..., IndSat5), we can see that Planning First is not able to solve the largest one, IndSat5. Planning First is faster than MAP-POP when dealing with small problems, but its performance decreases when the size of the problem increases. For instance, while the first three problems are solved faster by Planning First, it is slower than MAP-POP when solving IndSat4, and it does not find a solution to the most complex instance, IndSat5. MAP-POP proves also to be more effective at parallelizing actions in this domain as it obtains plans of equal or shorter duration than Planning First.

With respect to the \emph{rovers} domain, our results confirmed that MAP-POP solves 15 out of the 20 IPC problems for this domain. For the five IPC \emph{rovers} problems shown in Table \ref{Res_MAPSAP}, we can see the workload in this domain is better distributed than in the \emph{satellite} domain as most of the agents participate in the solution plan, which considerably reduces the duration of the plan. For instance, the solution plan for problem IPCRov7 contains 18 actions and is solved in just 6 time steps. Planning First solves only the two smallest IPC problems. For the \emph{independent} problems we modeled, Planning First obtains better-quality but more costly solutions than MAP-POP. The differences in execution time are far more noticeable than in the \emph{satellite} domain. This is due to to the more tightly-coupled nature of the problems of this domain (45.5$\%$ coupling level for the \emph{independent} problems), which affects negatively the performance of Planning First.

Finally, the \emph{logistics} domain has proven to be the most complex one for both multi-agent approaches. Agents in this domain are trucks and airplanes that must cooperate in most of the cases to transport packages. Hence, solutions for these problems are more costly than in the \emph{rovers} and \emph{satellite} domains, as they require agent coordination, an important feature to determine the efficiency of a MAP approach. Our results confirmed that MAP-POP loses performance in this domain, being able to solve only 9 out of 20 IPC problems. However, it distributes the workload effectively since all of the agents participate in all the solution plans obtained. Planning First shows also a poorer performance in this domain as it is not able to solve any of the IPC problems. These results are in line with the conclusions exposed in \cite{Nissim10}, which reveals the difficulty of a CSP-based approach to deal efficiently with problems that exhibit a high level of inter-agent interaction. As for the \emph{independent} problems, some of the solutions obtained by MAP-POP have better quality in terms of actions and duration than the solutions of Planning First. In addition, Planning First is still remarkably slower than MAP-POP, being unable to solve the IndLog5 problem, even though we defined rather small instances (notice the differences in execution time for the instance IndLog4). Again, Planning First only performs better than MAP-POP in the smaller problems.

The second test compares the scalability of both MAP frameworks, i.e. to which extent their efficiency is affected by the number of agents. In order to do so, we have run fourteen different tests for both the \textit{satellite} and the \textit{rovers} domains. Each test increases the number of agents in the task by one, from one agent to fourteen. The problems are modeled so that each of the participant agents has to achieve one of the problem's goals by itself.

Figure \ref{GraphSatellite} shows the scalability results for the \emph{satellite} domain. As it can be observed, Planning First show a better performance when solving small problems (up to seven agents). However, its performance decreases quickly as we execute larger problems. MAP-POP is faster at solving the 8-agent \emph{satellite} problem, and Planning First is unable to find a solution for the 9-agent problem upwards. MAP-POP, however, finds a solution for the 14 problem instances, and execution times suffer only a slight increase between problems.

The differences in performance of both systems are more noticeable in the more tightly-coupled \emph{rovers} domain. The results of this test are depicted in figure \ref{GraphRovers}. In this case, Planning First requires more than 40 minutes to solve the 6-agent \emph{rovers} problem, while MAP-POP takes only 20 seconds. Again, MAP-POP solves all the  problems without losing performance in the larger instances.

In conclusion, MAP-POP proves to be a more robust approach than Planning First as it can tackle larger and more complex planning problems. Moreover, while Planning First is designed for solving LCP, MAP-POP is a general-purpose method that tackles problems of different coupling levels. Although MAP-POP behaves better in LCP problems, it can also solve complex TCP problems. Scalability results show that Planning First performs better when dealing with simple problems that involve few agents. However, MAP-POP scales up far better, being able to solve much larger planning problems.

\section{CONCLUSIONS AND FUTURE WORK}

This paper presents a general-purpose MAP model suitable to cope with a wide variety of MA planning domains under incomplete information. The ability to define incomplete views of the world for the agents allows us to deal with more real problems, from inherently distributed domains -functionally or spatially- to problems that handle global and centralized sources of information. Currently, we are testing our planner on large-size logistics applications in which agents are geographically distributed and are completely unaware of the other agent's information except for the coordination points within their working areas.

The MAP resolution process is a POP-based refinement planning approach that iteratively combines planning and coordination while maintaining for each agent only the information that is visible to the planning entity.  This POP approach centered around the gradual construction of a joint solution plan for the MAP task highly benefits the resolution of cooperative distributed planning problems.

We have compared our MAP approach against Planning First, a system that handles agent coordination through a distributed CSP.  Results show that MAP-POP efficiently solves loosely-coupled problems but it also shows competitive when solving problems that have a higher coupling level and when computing plans that require the cooperation among agents. Hence, we can conclude that MAP-POP is an efficient, domain-independent and general-purpose framework to solve MAP problems.

\section*{ACKNOWLEDGEMENTS}

This work has been partly supported by the Spanish MICINN under projects Consolider Ingenio 2010 CSD2007-00022 and TIN2011-27652-C03-01, and the Valencian Prometeo project 2008/051.

\vspace{0.2cm}

\bibliography{biblio}
\bibliographystyle{ecai2012}

\end{document}